\documentclass[]{article}  

\usepackage[margin=1in]{geometry}
\usepackage{graphicx}
\usepackage{caption}
\usepackage{subcaption}
\usepackage{booktabs}
\usepackage{multirow}
\usepackage{amsmath} 
\usepackage{amssymb}  
\usepackage{csquotes}
\usepackage{enumitem}
\usepackage{url}
\usepackage{wrapfig}

\newcommand{\eg}{\emph{e.g.}, }
\newcommand{\ie}{\emph{i.e.}, } 

\title{\LARGE \bf An Analysis of Deep Object\\ Detectors for Diver Detection*}

\author{Karin de Langis$^{1}$, Michael Fulton$^{2}$ and Junaed Sattar$^{3}$
\thanks{The authors are with the Department of Computer Science and Engineering and the Minnesota Robotics Institute,
        University of Minnesota Twin Cities, Minneapolis, MN, USA.
        {\tt\small \{$^{1}$dento019, $^{2}$fulto081, $^{3}$junaed\}@umn.edu}}%
\thanks{*This work was supported by the US National Science Foundation Awards IIS-\#1845364 \& \#00074041 and the MnRI Seed Grant.}
}

\begin{document}

\maketitle

\begin{abstract}
With the end goal of selecting and using diver detection models to support human-robot collaboration capabilities such as diver following, we thoroughly analyze a large set of deep neural networks for diver detection. 
We begin by producing  a dataset of approximately 105,000 annotated images of divers sourced from videos -- one of the largest and most varied diver detection datasets ever created.
Using this dataset, we train a variety of state-of-the-art deep neural networks for object detection, including SSD with Mobilenet, Faster R-CNN, and YOLO.  
Along with these single-frame detectors, we also train networks designed for detection of objects in a video stream, using temporal information as well as single-frame image information.
We evaluate these networks on typical accuracy and efficiency metrics, as well as on the temporal stability of their detections.
Finally, we analyze the failures of these detectors, pointing out the most common scenarios of failure.
Based on our results, we recommend SSDs or Tiny-YOLOv4 for real-time applications on robots and recommend further investigation of video object detection methods.
\end{abstract}
\section{INTRODUCTION}
\label{sec:introduction}

Autonomous underwater vehicles (AUVs) are invaluable tools with great potential in advancing underwater science and engineering. 
AUVs already serve oceanographers and marine biologists by exploring environments and collecting data, and could one day aide in underwater construction and repair of infrastructure such as undersea internet cables, oil rigs, and much more.
Despite the fact that many of their advantages lie in their ability to replace humans or enter environments unsafe for humans, human-robot collaboration is still of interest in the AUV research space.
While there are interesting applications in which AUVs operate on their own, AUVs can also be seen as a useful tool to aide human workers underwater by carrying loads, recording data, or mapping environments while the human does other critical work. 
Rather than replacing human workers underwater, the advent of small, cheap, human-portable AUVs could be seen as an opportunity to improve the work of humans underwater by forming human-AUV teams.

A key capability of an AUV intended to work alongside humans is that of diver detection.
In order to safely move in environments with people in them, communicate with its operator, or follow them to a location, an AUV must have an understanding of where humans in its surroundings are. 
A common method for achieving this is the use of deep neural networks for object detection. 
Deep neural networks~\cite{goodfellow2016deep}, particularly single-frame convolutional neural networks (CNNs), have achieved great success in many object detection applications in the underwater space, including diver detection~\cite{islam_diver_2019}, coral identification \cite{ModasshirICRA2020}, fish species identification, and trash detection~\cite{Sattar2019ICRA-Fulton-Trash}.
However, not enough has been done to address the diver detection problem in terms of the practical considerations of deployment on robots.
In a robotic deployment, these detectors will be operating in the context of a video stream, with consequences to missed detections. 
Most previously presented diver detection systems achieve relatively high accuracies in the single image sense, but these metrics can hide a tendency for instability in detection bounding boxes when viewed in the temporal context.
The simplest applications of diver detection, diver following, can often make do with low-stability detections.
Other human-robot collaboration capabilities \enquote{down the pipeline}, however, such as gesture detection, attention and intention detection, and action recognition may not be able to cope with bounding boxes which frequently change in scale and position, or simply fail to detect for a particular frame.

 \begin{figure}[]
    \centering
    \vspace{2mm}
    \includegraphics[width=\textwidth, trim = 0cm 3cm 10cm 2cm, clip]{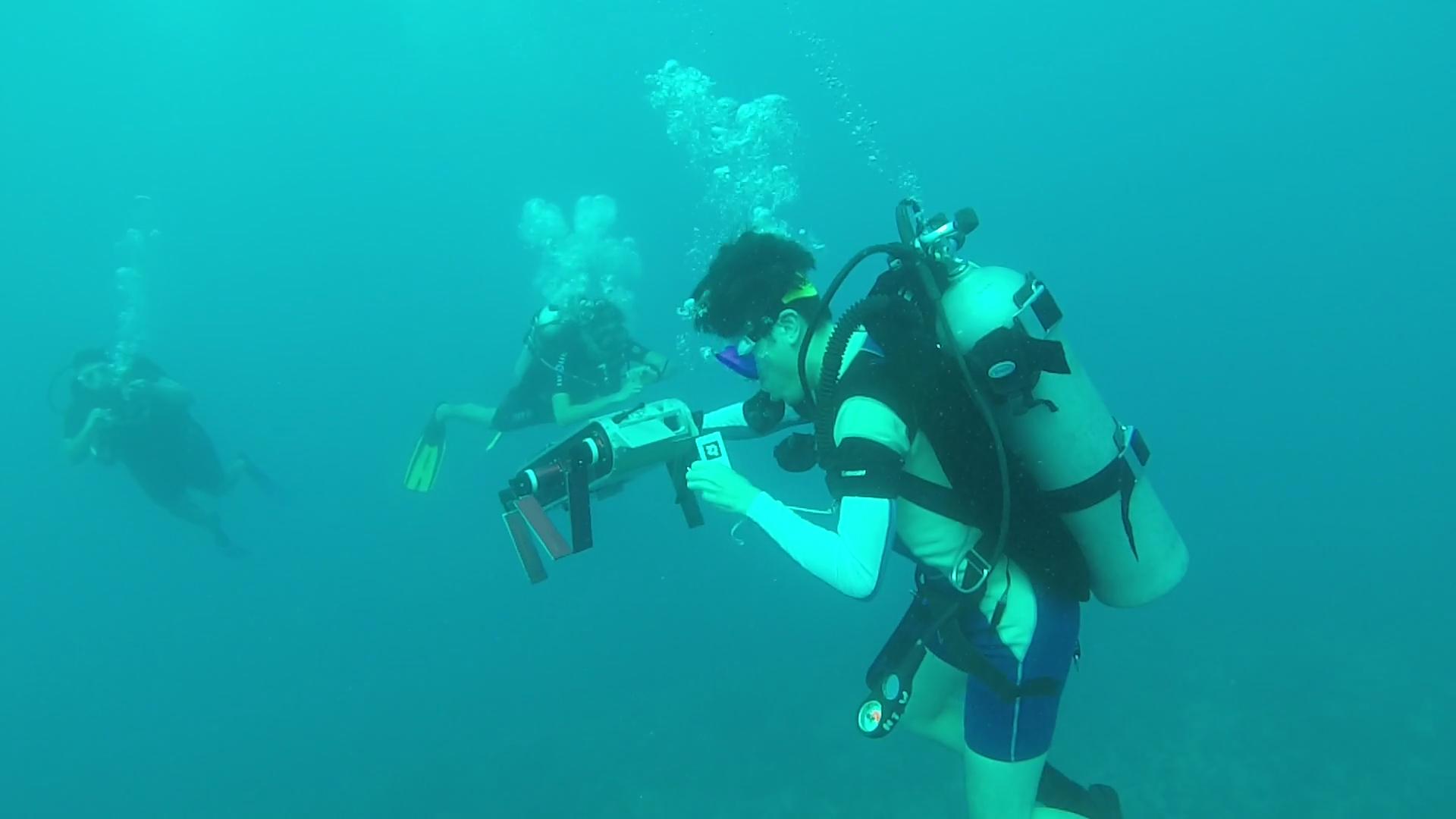}
    \caption{A diver operating an Aqua AUV in Barbados.}
    \label{fig:intro}
    \vspace{-3mm}
\end{figure}
To that end, we present an analysis of deep object detectors for diver detection. 
We begin by creating the Video Diver Detection dataset (VDD-$\bar{C}$\footnote{$\bar{C}$ is the Roman numeral for 100,000, denoting the number of images.}), a large dataset comprised of approximately $105,000$ fully annotated images of divers, drawn from videos taken in pool and field environments. 
This dataset improves on previous datasets in size (approximately 17 times more images as compared our previous work~\cite{islam_diver_2019}), but also by providing images in video context, allowing for analysis of the video-context stability of single-image detectors.
Additionally, video data is important for robotic applications: photographer bias in photo-based datasets means that video datasets have more realistic translations and rotations of divers that can encourage the network to be invariant to those transforms.
We then train a variety of single-image object detection networks on this dataset, as well as one video object detection network, and evaluate them all.
In order to build a full picture of their capabilities, we evaluate each network in terms of traditional accuracy metrics such as precision and recall, with a number of video stability metrics, and in terms of their inference capabilities on embedded platforms. 
The analysis presented in this paper explores the context of video object detection for the first time in the specific sub-field of diver detection, with the eventual goal of creating better diver detector systems.

\medskip

\noindent\textbf{Contributions}
In this paper we:
\begin{itemize}
    \item Create and process a $105,000$ image dataset of fully annotated videos of divers.
    \item Train four state-of-the-art deep learning-based diver detectors, each with a number of variants.
    \item Analyze these detectors in terms of accuracy, temporal stability, efficiency, and failure cases.
\end{itemize}
\section{RELATED WORK}
\label{sec:related}

Object detection is a computer vision task that involves identifying objects with a label and localizing objects with a bounding box. Convolutional neural networks (CNNs) are by far the highest preforming object detection models \cite{jiao2019survey, zhao2019object} and can generally be divided into two groups: two stage region-based detectors, which propose object regions in stage one and extract features from these regions in stage two (\eg Region CNN \cite{girshick2014rich} and its descendants Fast R-CNN \cite{girshick2015fast}, Faster R-CNN \cite{ren2015faster}, Region FCN \cite{dai2016r}, Mask R-CNN \cite{he2017mask}), and one stage grid-based detectors, which skip the region proposal step and instead extract features over a dense, static grid of possible object locations in the image (\eg SSD \cite{liu2016ssd}, YOLO \cite{redmon2016you}). One stage detectors are faster and suitable for realtime inference, while two stage detectors are more accurate but too slow for realtime applications.

State-of-the-art CNNs perform impressively well on vision benchmarks. However, these CNNs can stumble on images that come from a video stream \cite{azulay2019deep, shankar2019image, gu2019using, taori2020measuring}, which is particularly problematic for robotic applications \cite{sunderhauf2018limits}. Research investigating this phenomenon points to multiple reasons behind this performance deficit. One issue is that objects frequently move in a video, which means that objects  appear at a variety of locations. Image translations as small as one pixel can result in a radically different image representation at the deepest layers of state-of-the-art CNNs \cite{azulay2019deep}, which means that CNNs can struggle to generalize to the wide range of translations seen in video data. In fact, small translations of input images can be effective adversarial attacks on CNNs \cite{azulay2019deep, engstrom2019exploring, manfredi2020shift}. It is also important to note that CNNs are often trained on datasets like ImageNet \cite{deng2009imagenet} that have demonstrable location bias: the photographed objects' locations are not equally distributed throughout the dataset, and traditional data augmentation strategies do not sufficiently address this problem \cite{azulay2019deep, engstrom2019exploring}.

Similarly, objects in videos appear in a variety of orientations, which is also challenging for CNNs. Datasets typically present relatively head-on views of objects, whereas videos typically capture objects from a wide range of vantage points \cite{alcorn2019strike}. There is significant evidence that state-of-the art object detectors generalize very poorly to certain rotations \cite{engstrom2019exploring, alcorn2019strike}.

Learning to detect objects well in video is motivated by many applications, from robotics to surveillance. Since the release of ImageNet VID \cite{russakovsky2015imagenet} in 2015, researchers have developed many models for video object detection. These detectors can learn to exploit temporal information in video streams in order to make better detections on video data, and they typically outperform static image detectors on video datasets \cite{liu2018mobile}. Video detectors utilize a variety of strategies for leveraging temporal information, most notably linking static detections across frames in tracklets/tubelets \cite{feichtenhofer2017detect}, optical flow \cite{zhu2017flow}, and spatio-temporal feature memory \cite{bertasius2018object, liu2019looking, chen2020memory}. However, many video detectors cannot perform inference in real time, making them unsuitable for robotic applications. Most realtime-capable video detectors (e.g., \cite{liu2018mobile, zhu2018towards, liu2019looking}) achieve faster speeds by focusing intensive computational efforts on periodic ``key frames'' and propagating some of these computed features to subsequent frames, rather computing features for every single frame. 

Evaluation metrics are under-studied in video object detection. Video object detectors are typically evaluated with the same metrics used for static images detectors, \eg mean average precision (mAP). Notably, these metrics do not take into account the temporal nature of video data. Recently some metrics \cite{zhang2017stability, chen2020rethinking} have been proposed that evaluate video detectors not only on mAP, but also on the stability of bounding box location and scale for a given object across frames (\ie how much the bounding box jitters around the ground truth) and on how fragmented detections are for each object in the video (\ie during the duration of an object's presence in the video, how many times does an object's status change from ``detected'' to ``undetected''), although these metrics have not yet been widely adapted.


Diver detection is of significant interest to researchers in marine robotics, and previous work has collected various diver datasets for the purpose of diver detection. The Cognitive Autonomous Diving Buddy (CADDY) project is a broad collection of data for underwater vision research that includes a diver pose estimation dataset and a diver gesture recognition dataset \cite{gomez2019caddy, mivskovic2016caddy}. \cite{chavez2015visual} develop a diver detection algorithm that uses nearest-class-mean random forests. Our previous work develops a custom CNN for diver detection along with the Deep Diver Dataset (DDD) \cite{islam_diver_2019}. 
\section{DATASET CREATION AND PROCESSING}
\label{sec:dataset}
  \begin{figure}[]
    \centering
    \includegraphics[width=\textwidth]{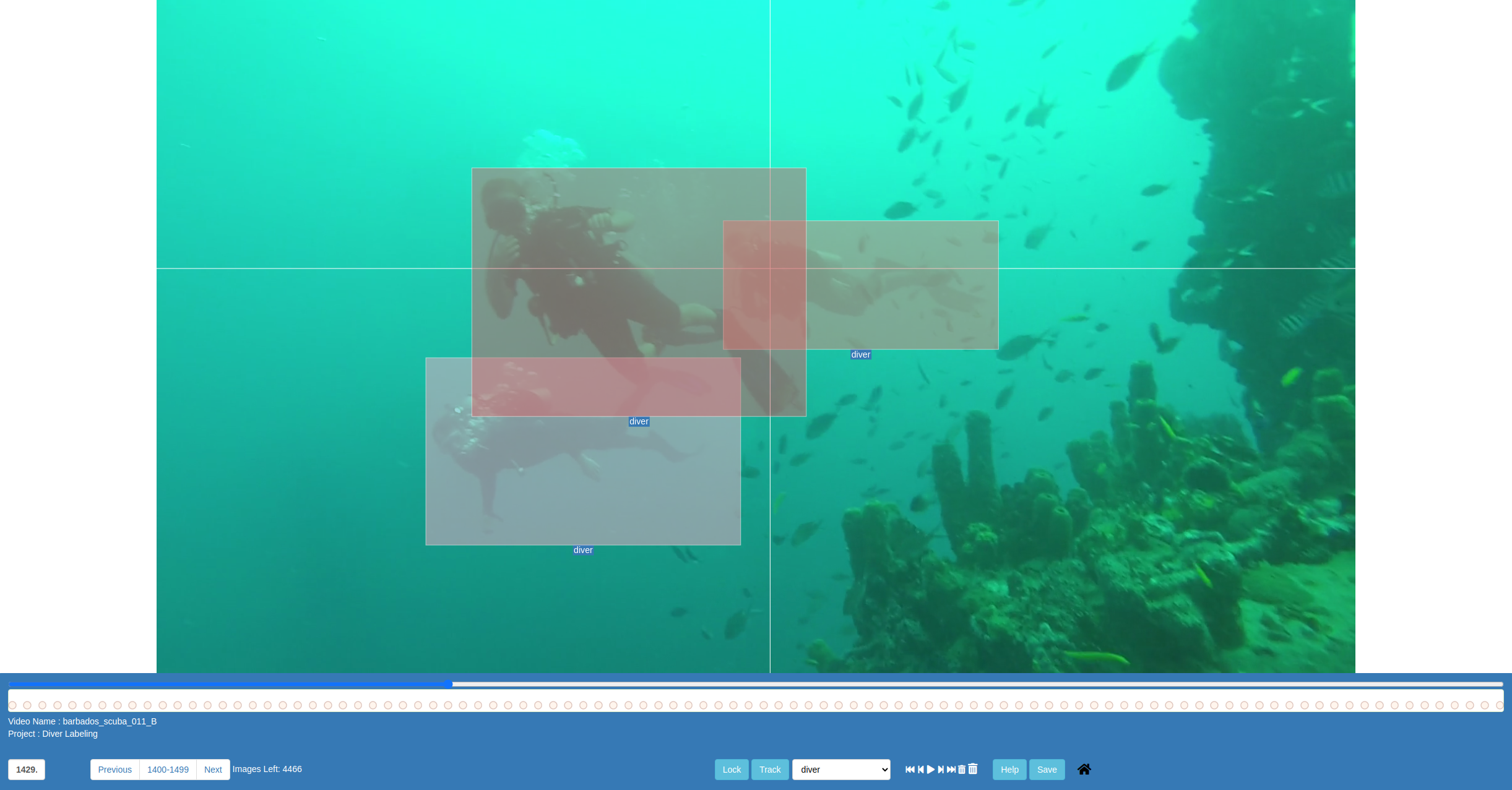}
    \caption{The EVA labeling tool.}
    \label{fig:eva}
    \vspace{-3mm}
\end{figure}

In order to train and evaluate deep neural networks for diver detection, we first need a dataset of properly annotated images. 
Our previous work produced a small dataset of images of divers in various environment~\cite{islam_diver_2019}.
This dataset will be referred to in this paper as the Deep Diver Dataset (DDD).
However, this dataset was deemed insufficient for the following reasons:
\begin{enumerate}[label=(\roman*)]
    \item It is a relatively small dataset from a deep learning perspective, with $6,011$ images in its training set.
    \item While many of the images are from videos, the organization of the dataset does not lend itself to temporal stability testing or training video detection methods.
    \item The majority of the training images are biased to the application of diver following: a single diver swimming away from the camera is a common image.
\end{enumerate}

To address these shortcomings, we present a new dataset, the Video Diver Detection dataset (VDD-$\bar{C}$).
We plan to release this dataset for public use in the near future.

\subsection{Source Data}
\label{sec:dataset:source}
\begin{figure}[]
    \vspace{2mm}
    \centering
    \begin{subfigure}[b]{0.32\textwidth}
        \centering
        \includegraphics[width=\textwidth, trim=0cm 3cm 0cm 0cm,clip]{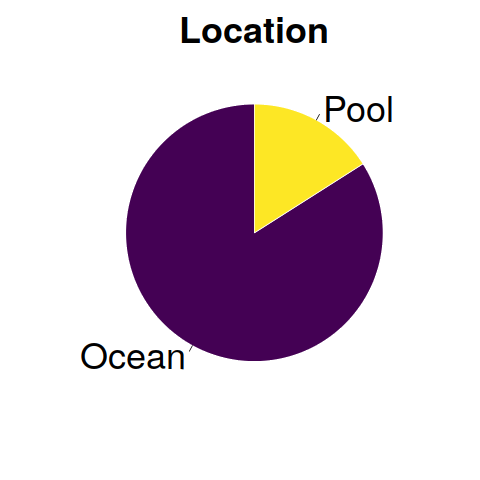}
        \caption{Video Location}
        \label{fig:vdd_dist:location}
    \end{subfigure}
    \hfill
    \begin{subfigure}[b]{0.32\textwidth}
        \centering
        \includegraphics[width=\textwidth, trim=0cm 3cm 0cm 0cm,clip]{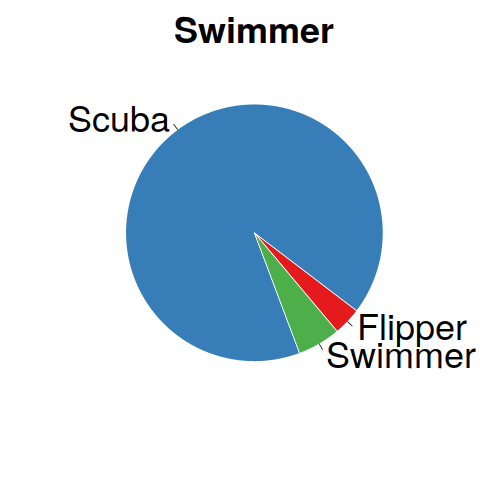}
        \caption{Type of swimmer}
        \label{fig:vdd_dist:swimmer}
    \end{subfigure}
        \begin{subfigure}[b]{0.32\textwidth}
        \centering
        \includegraphics[width=\textwidth, trim=0cm 3cm 0cm 0cm,clip]{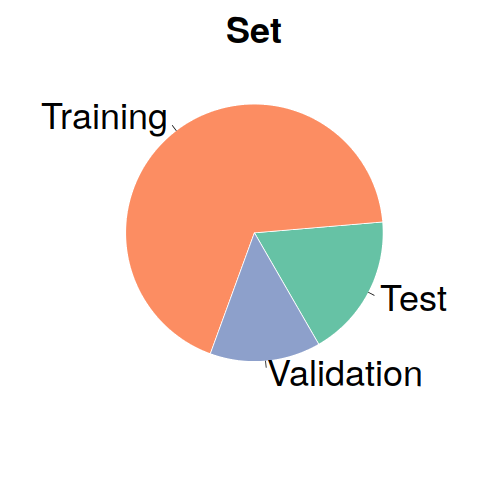}
        \caption{Train/Test/Val}
        \label{fig:vdd_dist:set}
    \end{subfigure}
        \begin{subfigure}[b]{0.32\textwidth}
        \centering
        \includegraphics[width=\textwidth, trim=0cm 3cm 0cm 0cm,clip]{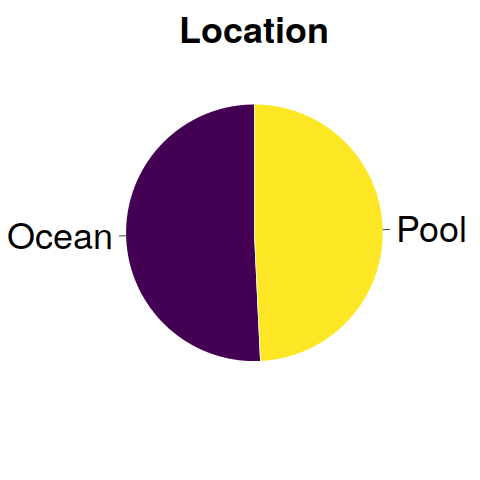}
        \caption{Video Location}
        \label{fig:ddd_dist:location}
    \end{subfigure}
    \hfill
    \begin{subfigure}[b]{0.32\textwidth}
        \centering
        \includegraphics[width=\textwidth, trim=0cm 3cm 0cm 0cm,clip]{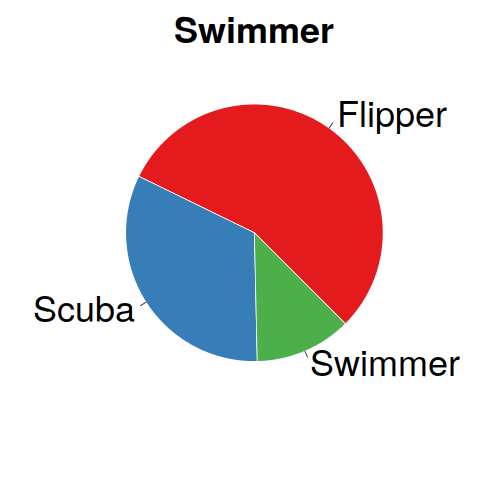}
        \caption{Type of swimmer}
        \label{fig:ddd_dist:swimmer}
    \end{subfigure}
        \begin{subfigure}[b]{0.32\textwidth}
        \centering
        \includegraphics[width=\textwidth, trim=0cm 3cm 0cm 0cm,clip]{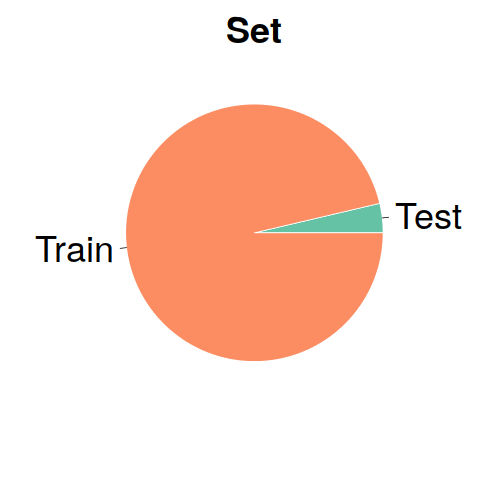}
        \caption{Train/Test/Val}
        \label{fig:ddd_dist:set}
    \end{subfigure}
    \caption{Distribution of VDD-$\bar{C}$ \textit{(a-c)} and DDD \textit{(d-f)} data.}
    \label{fig:dist}
    \vspace{-3mm}
\end{figure}

\begin{figure}[]
    \vspace{2mm}
    \centering
    \begin{subfigure}[b]{0.49\textwidth}
        \centering
        \includegraphics[width=\textwidth]{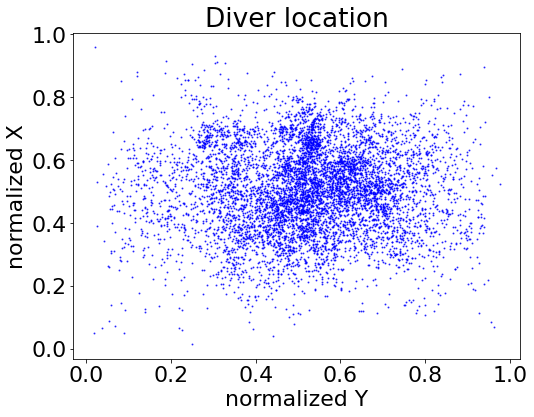}
        \caption{Deep Diver Dataset}
        \label{fig:centroids:ddd}
    \end{subfigure}
    \hfill
    \begin{subfigure}[b]{0.49\textwidth}
        \centering
        \includegraphics[width=\textwidth]{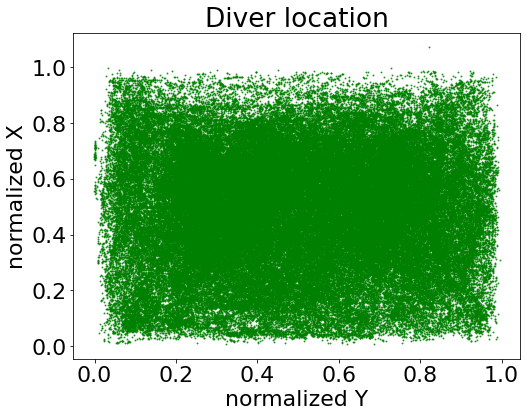}
        \caption{VDD-$\bar{C}$}
        \label{fig:centroids:vdd}
    \end{subfigure}
    \caption{Distribution of bounding box centers.}
    \label{fig:centroids}
\end{figure}

\begin{figure}[]
    \centering
    \begin{subfigure}[b]{0.49\textwidth}
        \centering
        \includegraphics[width=\textwidth]{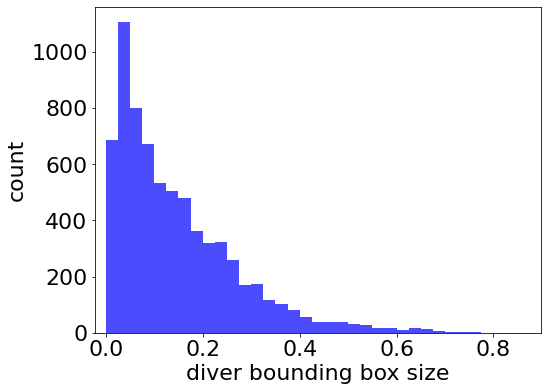}
        \caption{Deep Diver Dataset}
        \label{fig:scale:ddd}
    \end{subfigure}
    \hfill
    \begin{subfigure}[b]{0.49\textwidth}
        \centering
        \includegraphics[width=\textwidth]{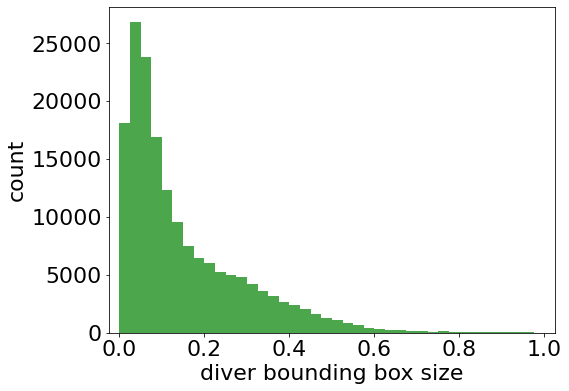}
        \caption{VDD-$\bar{C}$}
        \label{fig:scale:vdd}
    \end{subfigure}
    \caption{Distribution of bounding box areas.}
    \label{fig:scale}
    \vspace{-3mm}
\end{figure}

With the goal of temporal stability testing and video detection in mind, we chose to create our dataset out of videos, extracted into images at a rate of 20 frames per second for annotation.
The majority of the videos were from dives off the coast of Barbados in the Caribbean Sea, but a sizeable number of videos were taken in pool environments. 
The percentage of the dataset containing images from different environments (ocean/pool), images featuring different diver gear types (scuba/flippers/no gear), and images allocated to training, test, or validation sets is visualized in Figure \ref{fig:dist} for VDD-$\bar{C}$ (\ref{fig:vdd_dist:location} - \ref{fig:vdd_dist:set}) and DDD (\ref{fig:ddd_dist:location} - \ref{fig:ddd_dist:set}).
Note that the the Figure shows percentages rather than number of frames.
For instance, while a smaller percentage of VDD-$\bar{C}$'s total data is from pool environments, it has nearly three times as many images of pool environments ($16,657$) as DDD has images of any type.
Additionally, while there is less variety in what equipment divers are wearing, a much wider array of viewpoints are represented: divers were recorded swimming with or without a robot, viewed from many different angles, and sometimes merely treading water.
Analysis of bounding box centroid location (Figure \ref{fig:centroids}) and bounding box area (Figure \ref{fig:scale}) clearly shows the greater variety in diver locations present in our new dataset.
In comparison to previous datasets, our source data is by far more numerous, contains a sufficient amount of diver and environment variation, and has a much wider range of viewpoints and diver activities represented.

\subsection{Labeling Process}
\label{sec:dataset:labeling}
\begin{figure}[]
    \vspace{2mm}
    \centering
    \begin{subfigure}[b]{0.49\textwidth}
        \centering
        \includegraphics[width=\textwidth]{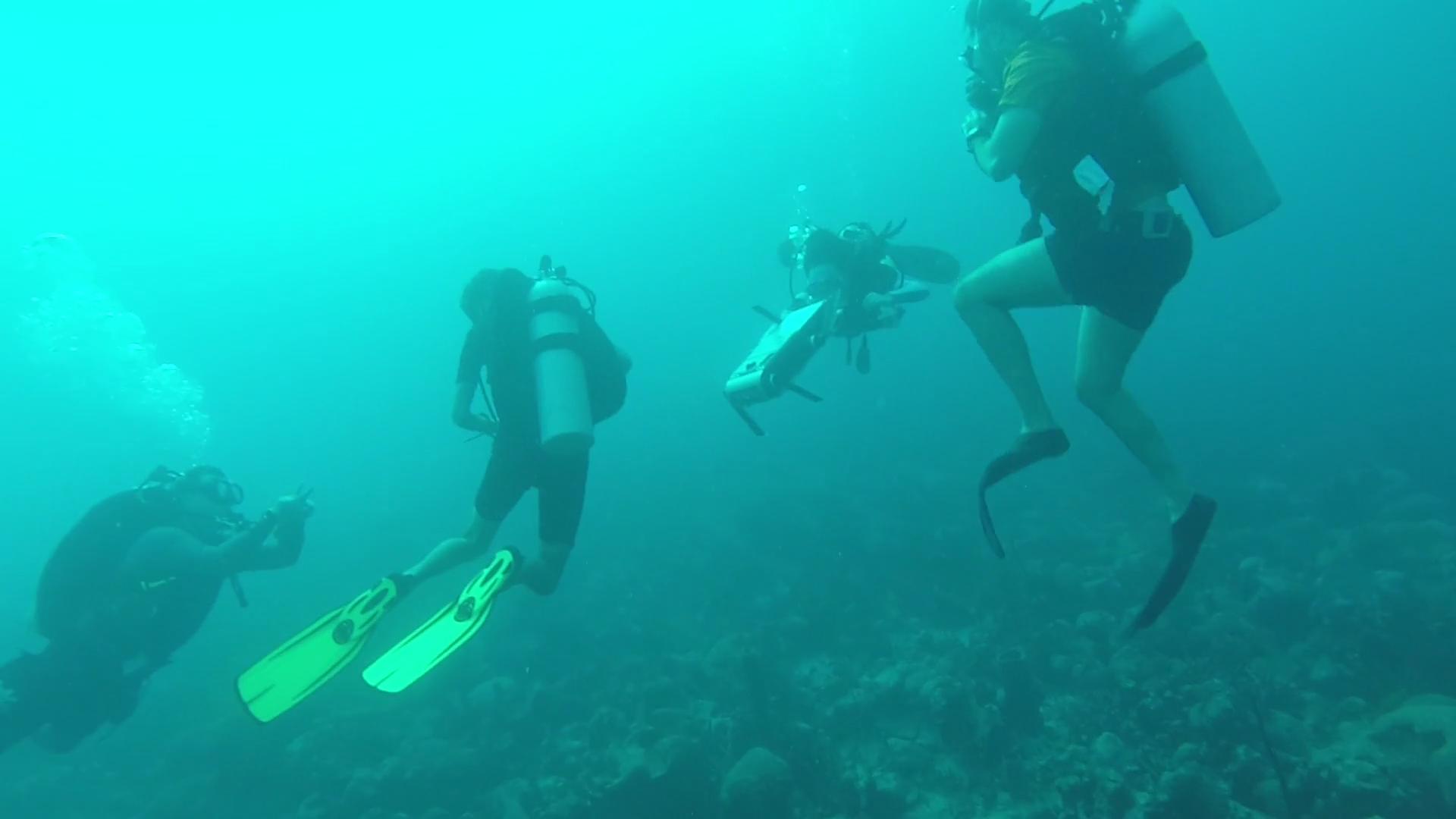}
        \caption{Unlabeled}
        \label{fig:dataset:unlabled}
    \end{subfigure}
    \hfill
    \begin{subfigure}[b]{0.49\textwidth}
        \centering
        \includegraphics[width=\textwidth]{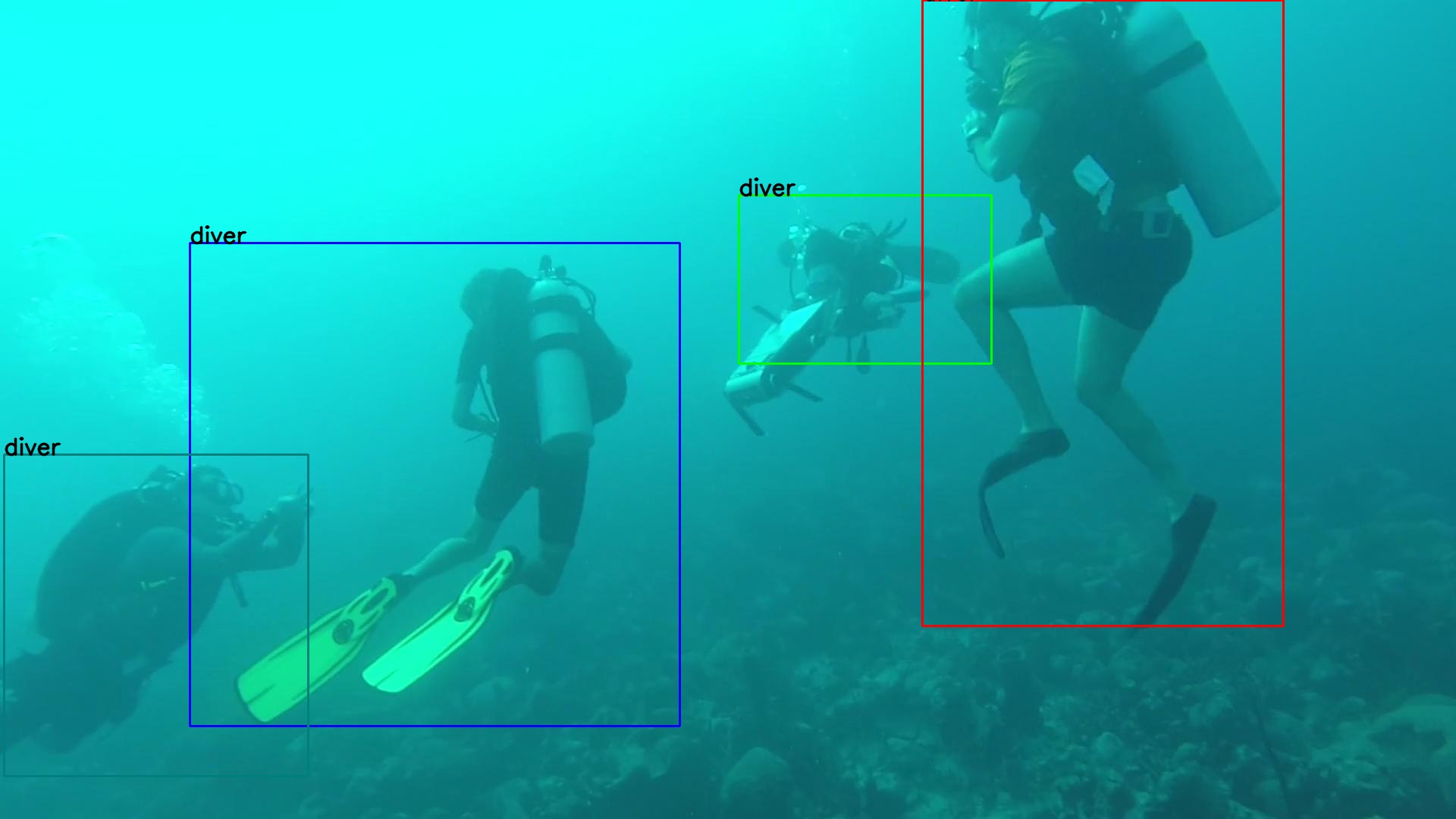}
        \caption{Labeled}
        \label{fig:dataset:labled}
    \end{subfigure}
    \caption{An image from VDD-$\bar{C}$, with and without labels.}
    \label{fig:dataset}
    \vspace{-3mm}
\end{figure}

Once the videos for the dataset had been selected and extracted to frames, the task of labeling them was addressed. 
Labeling $105,000$ images one by one was an extremely difficult and time-consuming task, but it was improved by our choice of labeling tool.
We used EVA~\cite{ericsson_eva}, a web-based tool for labeling video data, shown in Figure \ref{fig:eva}. 
EVA is a rebuild of the popular Beaver-Dam tool, with the addition of tracking capabilities. 
Annotation is completed normally for the first frame in a video, with a user drawing a bounding box around every object they wish to label (Figure~\ref{fig:dataset}). 
Then, the user clicks the track button, and the initial annotations are used to initialize a Kernelized Correlation Filter tracker~\cite{henriques_2012_kcf} which propagates those bounding boxes over the following frames. 
Depending on the difficulty of the tracking, the generated bounding boxes need to be adjusted and re-tracked somewhere between every frame and every 30 frames.

\subsection{Post-Processing}
\label{sec:dataset:processing}
With the initial labels generated, we began post-processing our data, beginning with a significant proofreading effort, in which every labeled video was watched from start to finish, with all labeling errors noted down. 
Following the correction of these errors, a number of sections of video were cut due to significant blurring or degradation of the visual quality. 
Additionally, any frame in a pool video which contained no diver was cut, as these frames were almost entirely from portions of the video with the camera out of water.
A significant number of images were cut, bringing the total number of images down from approximately $114,000$ to the $105,000$ images previously mentioned.
Finally, the exported Pascal VOC annotations were automatically filtered for bounding box coordinates out of the acceptable range of the image size before being converted to YOLO~\cite{redmon_2016_yolo}-style labels, TFExample~\cite{tf_example} and TFSequence~\cite{tf_sequence} records. 

\section{DETECTION MODELS}
\label{sec:models}

With the dataset constructed, we turn our attention to the deep detection models that will be trained using it.
We trained four models (some with a few variants) of two different types.
In the following sections we briefly explain these models, along with the training process we used for each, and any variants for which we report results.

\subsection{Faster R-CNN}
\label{sec:models:frcnn}
Faster R-CNN \cite{ren2015faster} is a two-stage object detector in the R-CNN family and a staple high accuracy object detector. Although it is not fast enough for use on a robotic platform, we chose to train this model to loosely represent a ``top end" accuracy of state-of-the-art CNNs on our dataset. We utilize the Tensorflow Object Detection API \cite{huang2017speed} and its Faster R-CNN with an Inception-ResNet-v2 \cite{szegedy2017inception} feature extractor for training. Two hyperparameters, learning rate and batch size, are tuned with the validation set.

\subsection{SSD with Mobilenet}
\label{sec:models:ssd}
SSDs \cite{liu2016ssd} are among the most accurate real-time object detectors and therefore are good candidates for eventual deployment on a robotic platform. We train SSD320 (\ie SSDs for inputs sized $320 \times 320$) models with multiple Mobilenet \cite{howard2017mobilenets} backbones to find the optimal model for our use case. We utilize the Tensorflow Object Detection API and the provided models for training. Learning rate and batch size are again tuned with the validation set.

\subsection{YOLO}
\label{sec:models:yolo}
You Only Look Once (YOLO)~\cite{redmon_2016_yolo} is at this point a well established object detection model, valued for its high accuracy and speed. 
YOLO predicts a set of bounding boxes in a grid across the image, with confidences for each, and class probabilities for each grid box, matching class probabilities to the boxes with the highest confidences.
We evaluate a variety of versions of YOLO (v2~\cite{redmon_2017_yolov2} and v4~\cite{bochkovskiy_2020_yolov4}) in this work, although YOLOv4 is our primary version for comparing with other networks. 
For every version of YOLO we train, we also train Tiny-YOLO, which reduces the number of convolutional layers and filters to improve the inference runtime of the network.
To train these networks, we fine-tune them using initial weights trained on Imagenet.

\subsection{LSTM-SSD}
\label{sec:models:yolo:conv-lstm}
The only video object detection network evaluated in this work is the LSTM-SSD detector proposed by Liu and Zhu~\cite{liu2018mobile}.
This model is based on Mobilenet SSDs, but adds a number of Bottleneck LSTMs~\cite{liu2018mobile} after the feature extraction network, followed by output layers.
On the next frame's inference, features extracted by the convolutional layers will be combined with the LSTM's state, propagating feature maps through time.
In order to train the LSTM-SSD, we initialize the convolutional portion of the network from a fine-tuned MobileNetV1 SSD, then train the LSTM portion of the network in order to improve the feature propagation.

\section{Temporal Stability}
\label{sec:stability}

\begin{table*}[]
    \vspace{2mm}
    \centering
    \resizebox{\textwidth}{!}{%
    \begin{tabular}{lccccccccccccc}
    & \multicolumn{6}{c}{\textbf{Trained on VDD-$\bar{C}$}} && \multicolumn{6}{c}{\textbf{Trained on DDD}}\\
    \cmidrule[0.9pt](l){2-7} \cmidrule[0.9pt](l){9-14}
    
    & \multicolumn{3}{c}{\textbf{Tested on VDD-$\bar{C}$}} & \multicolumn{3}{c}{\textbf{Tested on DDD}} && \multicolumn{3}{c}{\textbf{Tested on VDD-$\bar{C}$}} & \multicolumn{3}{c}{\textbf{Tested on DDD}}\\
    
    \cmidrule[0.9pt](l){2-4}\cmidrule[0.9pt](l){5-7}\cmidrule[0.9pt](l){9-11}\cmidrule[0.9pt](l){12-14}
    
    \textbf{Model} & \textbf{AP$_{50}$} & \textbf{AP$_{75}$} & \textbf{IOU} & \textbf{AP$_{50}$} & \textbf{AP$_{75}$} & \textbf{IOU} && \textbf{AP$_{50}$} & \textbf{AP$_{75}$} & \textbf{IOU} & \textbf{AP$_{50}$} & \textbf{AP$_{75}$} & \textbf{IOU} \\
    \cmidrule[0.9pt](l){1-14}
    
    SSD(MobileNetV2) & 82.43 & 41.07 & 39.61 & 90.12 & \textbf{33.09} & 65.14 && 69.14 & 25.61  & 45.73 & 85.90 & 19.8 & 65.80 \\
    SSD(MobileNetV3-Small) & 81.43 & 34.03 & 29.50 & 60.20 & 11.10 & 58.50 && 60.24 & 11.10 & 31.23 & 83.90 & 17.46 & 65.80 \\
    SSD(MobileNetV3-Large) & \textbf{88.47} & \textbf{44.16} & 40.42 & 91.29 & 27.94 & 66.21 && \textbf{77.01} & \textbf{26.19} & 40.01 & 89.81 & \textbf{30.20} & 65.85 \\
    YOLOv2  & 86.73	& 38.04	& \textbf{68.25} & 93.30 & 23.89 & 68.19 && 76.27 & 19.3 & \textbf{60.92} & 87.84 &	21.93 &	68.76 \\
    YOLOv2-Tiny  & 72.69 & 8.82 & 59.48 & 63.50 & 3.14& 60.12 && 66.51 & 6.17 & 55.47 & 81.95 & 3.27 & 63.95\\
    YOLOv4  & 83.65 & 34.13 & 64.16 & 81.38 & 21.64 & \textbf{71.13}	&& 70.47 & 20.86 & 59.25 & \textbf{91.57} & 18.23 & 71.13\\
    YOLOv4-Tiny &  81.93 & 34.7 & 58.82 & \textbf{92.15} & 26.26 & 68.00 && 75.56 & 19.43 & 57.83 & 84.41 &11.44 &	\textbf{73.56} \\
    
    \end{tabular}
    }
    \caption{Comparison between performance on test sets of the VDD-$\bar{C}$ and DDD with different training sets.}
    \label{tab:dataset_comparison}
    \vspace{-5mm}
\end{table*}

In robotic applications, it is often desirable to have temporal stability for object detections. That is, detections for a given object should be stable over time with respect to: \begin{itemize}
    \item \textit{Translation.} If detected bounding boxes are not consistently located with respect to the ground truth bounding boxes from frame to frame, it is difficult to estimate the object's location and trajectory.
    \item \textit{Scale and aspect ratio.} Similarly, if detected bounding boxes have inconsistent scales and aspect ratios, it is difficult to estimate the object's location and trajectory.
    \item \textit{Fragmentation.} For any given diver that appears in the video, the diver should be consistently detected (\ie the object should not be undetected in one frame, then detected in the next, and so on). At worst, fragmentations make it difficult for the robot to confidently determine that the object is present, and at best, they increase uncertainty of estimations of the object's location.   
\end{itemize}
We adapt methods from \cite{zhang2017stability} and \cite{chen2020rethinking} to evaluate diver detectors with respect to these three aspects of temporal stability. As in \cite{chen2020rethinking}, we do not have ground truth tracks for the divers in our dataset and therefore computationally calculate ad-hoc tracklets for each diver by matching ground truth annotations from frame to frame with intersection over union (IOU) as in \cite{chen2019temporally}. Using these tracklets, we then calculate the stability metrics from \cite{zhang2017stability} as follows:
\subsubsection{Translation error}
The translation error of each tracklet's detection is measured with the center position error $e_c$. To calculate $e_c$, for each detection $d$ in the tracklet, we find the standard deviation of the distance between the normalized center $x$ bounding box coordinate, $x_d$,  and the normalized ground truth $x_g$. We do the same for the $y_d$ and $y_g$ coordinates. The translation error is the mean of these standard deviations across all tracks. Formally, for each tracklet $t$:
\begin{align*}
e_c(t) = \sigma(x_d - x_g) + \sigma(y_d - y_g), \forall d \in t
\end{align*}
Then the detector's overall translation error is
\begin{align*}
\frac{1}{N}\sum_{t = 1}^{N}e_c(t)
\end{align*}
\subsubsection{Scale and aspect ratio error}
For each detection, the aspect ratio error is defined as the ratio between the bounding box aspect ratio and the ground truth aspect ratio. The scale error is defined as the square root of the bounding box area over the ground truth area. To find the scale and aspect ratio error, we find the average standard deviations of each track's summed scale error $e_s(t)$ and aspect ratio error $e_r(t)$. Formally, for each tracklet $t$:
\begin{align*}
e_s(t) &= \sigma\left(\sqrt{\frac{w_dh_d}{w_gh_g}}\right), \forall d \in t\\
e_r(t) &= \sigma\left(\frac{w_d}{h_d}/\frac{w_g}{h_g}\right), \forall d \in t \\
e_{sr}(t) &= e_s(t) + e_r(t) 
\end{align*}
Then the detector's overall scale and aspect ratio error is
\begin{align*}
\frac{1}{N}\sum_{t = 1}^{N}e_{sr}(t)
\end{align*}
\subsubsection{Fragmentation error}
For each track, we count the number of fragments as the number of times the track's status changes from detected to undetected or vice versa. Then the fragmentation error is the average number of fragments $f$ per track, normalized by track length $l$:
\begin{align*}
    \frac{1}{N}\sum_{t = 1}^{N}\frac{f_t}{l_t -1}
\end{align*}
Because the translation and scale metrics rely on standard deviations of a tracklet's detections, they become meaningless for tracklets with only one detection. Therefore in our analysis we exclude any tracklets with only one matched detection.

\section{RESULTS}
\label{sec:results}

\subsection{Dataset Comparisons}
\label{sec:results:dataset}
To quantify how effective the new VDD-$\bar{C}$ dataset is in training deep vision models compared to previous datsets, we train one version of a group of common models on the VDD-$\bar{C}$ dataset and a second version on the existing DDD dataset. 
The models trained are versions of SSD and YOLO -- we do not train Faster R-CNN or LSTM-SSD on DDD, so we cannot compare their performance.
We then compare their respective performances on each test set as shown in Table \ref{tab:dataset_comparison}.
Results show that models trained on VDD-$\bar{C}$ outperform those trained on DDD on both the VDD-$\bar{C}$ test set and, to a lesser extent, the DDD test set.
These results support our expectations that VDD-$\bar{C}$'s more complex data will lead to more successful detectors, because the VDD-$\bar{C}$ trained models outperform the DDD-trained models with few exceptions. 
Additionally, our hope that the challenge presented by our test dataset would be greater is reflected in these results, as DDD-trained detectors perform more poorly on our test data than the DDD test set.

\begin{wraptable}[11]{l}{0.5\textwidth}
    \centering
    \setlength\tabcolsep{3pt}
    \resizebox{.45\textwidth}{!}{%
    \begin{tabular}{lccccc}
         \textbf{Model} & \textbf{AP} & \textbf{AP$_{50}$} & \textbf{AP$_{75}$} & \textbf{IOU} \\
         \cmidrule[0.9pt](l){1-6}
         Faster R-CNN & 55.50 & 90.18 & 60.50 & 49.81 \\
         \cmidrule[0.9pt](l){2-5}
         SSD(MobileNetv2) & 43.45 & 82.43 & 41.07 & 39.61 \\
         SSD(MobileNetv3-Small) & 39.81 &  81.43 & 34.03 &  29.50 \\
         SSD(MobileNetv3-Large) & \textbf{47.05} & \textbf{88.47} & \textbf{44.16} & 40.42  \\
         YOLOv4 & 41.01 & 83.65 & 34.13 & \textbf{64.16} \\
         YOLOv4-Tiny & 33.39 & 81.93 & 34.70 & 58.81 \\
         LSTM-SSD & 39.00 & 79.80 & 33.10 & 51.40 &  \\
    \end{tabular}
    }
    \caption{Precision and IOU.}
    \label{tab:precision_results}
\end{wraptable}

\subsection{Average Precision and IOU}
\label{sec:results:model:pri}
To evaluate the accuracy of each model, we calculate the average precision (AP) of each model on diver identification. 
The average precision is found by evaluating the model's precision at different recall values. Specifically, since models output a confidence score for each detection, model recall values can be manipulated by changing the confidence threshold required for a detection; the AP is the weighted mean of the model's precision values at each recall value, where the weight for the recall at a given confidence threshold is the increase in recall from the previous threshold. Note that since our models are only trained to identify divers, the diver AP is equivalent to the mean average precision (mAP), which is a widely used object detection metric \cite{zhao2019object}.
For each model, we pick a confidence threshold that results in the best precision and recall scores. 
Using that confidence threshold, we calculate AP at IOU thresholds of 0.5 and 0.75, average IOU, and an average of APs with thresholds between 0.5 and 0.95 with a step size of 0.05 (0.5-0.95).

\subsection{Stability Results}
\label{sec:results:model:stability}
The average translation, scale, and fragment errors across all diver tracklets are calculated for each model using the equations discussed in Section \ref{sec:stability}. The models perform very similarly with respect to translation error and scale error. Fragmentation error varies more across models. The YOLOv4 and YOLOv4-tiny models are the most accurate and have the lowest fragmentation errors. Notably, while the LSTM-SSD and SSD-Mobilenets have comparable accuracy, the LSTM-SSD has a lower fragmentation error, indicating that it outperforms SSD in detecting divers consistently.

\begin{figure}[]
    \centering
    \includegraphics[width=0.75\textwidth]{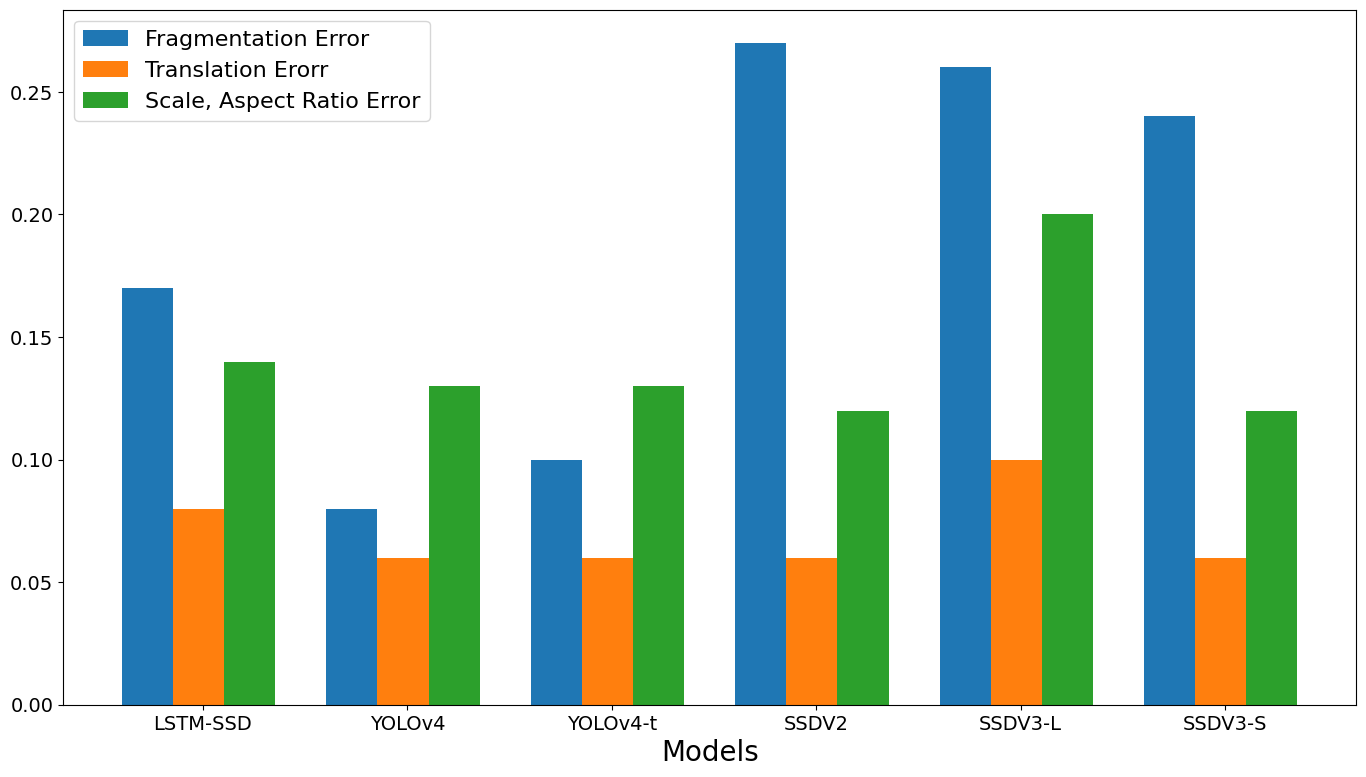}
    \caption{Measured stability errors.}
    \label{fig:stability_graph}
\end{figure}

\begin{wraptable}[11]{l}{0.5\textwidth}
    \centering
    \setlength\tabcolsep{3pt} 
    \resizebox{.45\textwidth}{!}{%
    \begin{tabular}{lcc}
         \textbf{Model} & \textbf{FPS(GPU)} & \textbf{FPS(TX2)} \\
         \cmidrule[0.9pt](l){1-3}
         SSD(MobileNetv2) & 50 & 9  \\
         SSD(MobileNetv3-Small) & 52 & 9 \\
         SSD(MobileNetv3-Large) & 51 & 8 \\
         YOLOv4 & 50 & 5    \\
         YOLOv4-Tiny & 88 & 35 \\
         LSTM-SSD & 73  & 19  \\
    \end{tabular}
    }
    \caption{Frames per second for inference.}
    \label{tab:efficiency_results}
\end{wraptable}

\subsection{Efficiency Results}
\label{sec:results:model:efficiency}
YOLO, SSD, and LSTM-SSD are high speed models designed for real-time inference use cases.
While Faster R-CNN has demonstrated high accuracies, it is computationally involved, and is not quite suitable for real-time inference as shown in ~\cite{islam_diver_2019}.
In order to quantify the usability of our real-time models on robotic platforms, we quantify their inference run-time in terms of frames processed per second (FPS).
The results of these tests can be seen in Table \ref{tab:efficiency_results}.
Due to the size of the test dataset, we only tested a portion of the test set for runtime calculation: 5,000 randomly selected frames.
We tested each network on two devices: an Nvidia 1080 GPU and an Nvidia Jetson TX2. 
These results do not represent the maximum inference speed possible, as no platform-specific optimization was done, but they provide a guide to the applicability of these networks in embedded contexts, on board AUVs. 
While the SSD variants achieve relatively high framerates on embedded devices, the clear standout is YOLOv4-Tiny, which achieves real-time performance with accuracy near that other methods. 
LSTM-SSD also performs quite well, surpassing the traditional SSD variants.

\subsection{Failure Scenarios}
\label{sec:results:failures}
\begin{figure}[]
    \centering
    \includegraphics[width=0.75\textwidth]{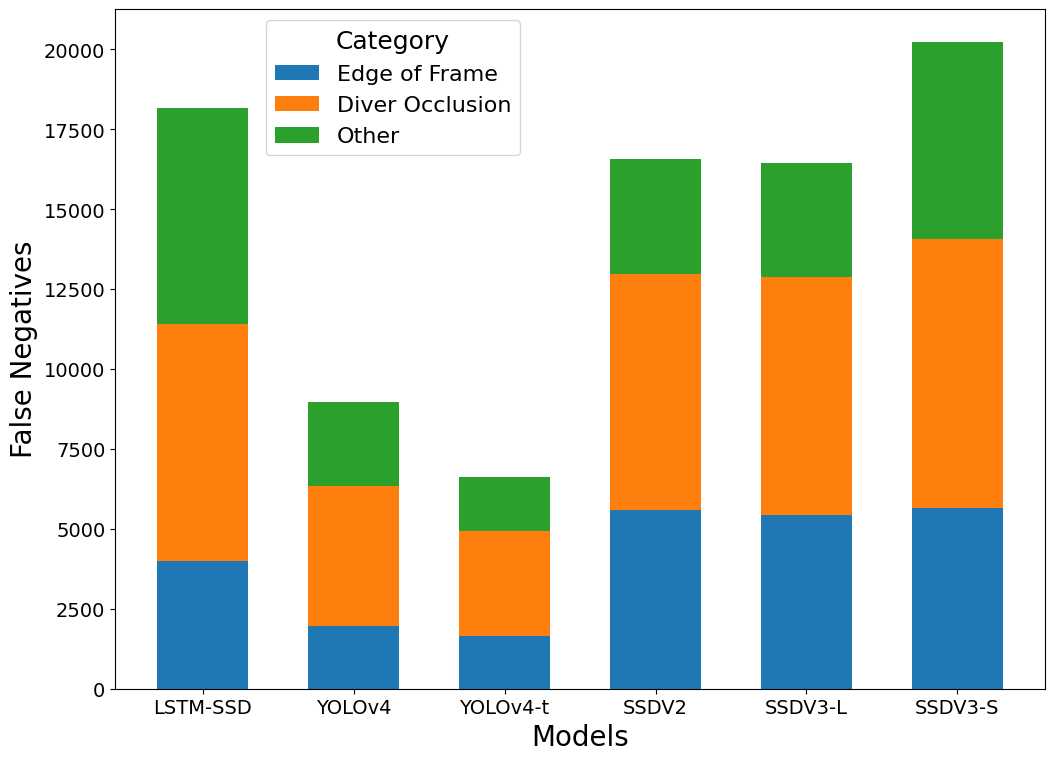}
    \caption{The source of false negative errors.}
    \label{fig:fn_graph}
\end{figure}

When considering a diver detector for use on an AUV, there is information of interest beyond accuracy, stability, and efficiency: when and why the detector fails.
By inspecting the instances of false negative detections, we can gain some intuition on the circumstances of detector failures in the diver detector task.
A significant portion of false negatives stem from one of two cases: divers not fully in the frame, or diver occlusions, as shown in Figure \ref{fig:fn_graph}. We define a diver as not fully in the frame if an edge of their ground truth bounding box is on the edge of the frame. We define a diver occlusion as two ground truth bounding boxes with an IOU above 0. 

\section{Conclusion}
\label{sec:conclusion}
In this paper we presented VDD-$\bar{C}$, a large dataset of annotated videos of divers for use in training diver detectors.
With this dataset, we trained a variety of detectors: a two-stage network (Faster R-CNN), single stage networks (SSD and YOLO), and a video object detector (LSTM-SSD).
We showed that our dataset improved the quality of detection results significantly, while simultaneously providing a stronger challenge in the test set. 
In terms of precision and efficiency results we reproduced previous work, showing that SSD and YOLO networks have similar accuracies and mobile inference runtimes, though LSTM-SSD outpaces all but Tiny-YOLO for mobile inference.
Our stability testing revealed some benefit to the video object detection methods of LSTM-SSD over traditional SSDs when it comes to fragmentation error, though YOLO networks performed even better on that metric.
Finally, analysis of failure cases revealed divers not fully in the frame and diver-on-diver occlusion as the greatest sources of failures. 
Our recommendations for those who are searching for a diver detection solution, based on these results, is that users select Tiny-YOLO if real-time inference is of utmost importance or an SSD model if more consistent accuracy is required at the cost of inference time.
Lastly, we recommend further exploration of video object detection methods for future work. 
While LSTM-SSD did not preform sufficiently on accuracy metrics for our uses, its low inference times and high stability reveal the potential benefits of video object detection in this application.

\bibliographystyle{abbrv}
\bibliography{references.bib}

\begin{thebibliography}{10}

\bibitem{alcorn2019strike}
M.~A. Alcorn, Q.~Li, Z.~Gong, C.~Wang, L.~Mai, W.-S. Ku, and A.~Nguyen.
\newblock Strike (with) a pose: Neural networks are easily fooled by strange
  poses of familiar objects.
\newblock In {\em Proceedings of the IEEE Conference on Computer Vision and
  Pattern Recognition (CVPR)}, pages 4845--4854, 2019.

\bibitem{azulay2019deep}
A.~Azulay and Y.~Weiss.
\newblock Why do deep convolutional networks generalize so poorly to small
  image transformations?
\newblock {\em Journal of Machine Learning Research}, 20(184):1--25, 2019.

\bibitem{bertasius2018object}
G.~Bertasius, L.~Torresani, and J.~Shi.
\newblock Object detection in video with spatiotemporal sampling networks.
\newblock In {\em Proceedings of the European Conference on Computer Vision
  (ECCV)}, pages 331--346, 2018.

\bibitem{bochkovskiy_2020_yolov4}
A.~Bochkovskiy, C.-Y. Wang, and H.-Y.~M. Liao.
\newblock Yolov4: Optimal speed and accuracy of object detection.
\newblock {\em arXiv preprint ArXiv:2004.10934}, 2020.

\bibitem{chavez2015visual}
A.~G. Chavez, M.~Pfingsthorn, A.~Birk, I.~Renduli{\'c}, and N.~Miskovi{\'c}.
\newblock Visual diver detection using multi-descriptor nearest-class-mean
  random forests in the context of underwater human robot interaction (hri).
\newblock In {\em OCEANS}, pages 1--7. IEEE, 2015.

\bibitem{chen2020rethinking}
X.~Chen, Z.~Wu, J.~Yu, and L.~Wen.
\newblock Rethinking temporal object detection from robotic perspectives.
\newblock {\em arXiv preprint arXiv:1912.10406}, 2020.

\bibitem{chen2019temporally}
X.~Chen, J.~Yu, and Z.~Wu.
\newblock Temporally identity-aware {SSD} with attentional {LSTM}.
\newblock {\em IEEE Transactions on Cybernetics}, 50(6):2674--2686, 2019.

\bibitem{chen2020memory}
Y.~Chen, Y.~Cao, H.~Hu, and L.~Wang.
\newblock Memory enhanced global-local aggregation for video object detection.
\newblock In {\em Proceedings of the IEEE/CVF Conference on Computer Vision and
  Pattern Recognition (CVPR)}, pages 10337--10346, 2020.

\bibitem{dai2016r}
J.~Dai, Y.~Li, K.~He, and J.~Sun.
\newblock {R-FCN}: Object detection via region-based fully convolutional
  networks.
\newblock In {\em Advances in Neural Information Processing Systems (NIPS)},
  pages 379--387, 2016.

\bibitem{deng2009imagenet}
J.~Deng, W.~Dong, R.~Socher, L.-J. Li, K.~Li, and L.~Fei-Fei.
\newblock Imagenet: A large-scale hierarchical image database.
\newblock In {\em 2009 IEEE conference on computer vision and pattern
  recognition}, pages 248--255. Ieee, 2009.

\bibitem{engstrom2019exploring}
L.~Engstrom, B.~Tran, D.~Tsipras, L.~Schmidt, and A.~Madry.
\newblock Exploring the landscape of spatial robustness.
\newblock In {\em International Conference on Machine Learning}, pages
  1802--1811, 2019.

\bibitem{ericsson_eva}
{Ericsson Open Source Software}.
\newblock Github - ericsson/eva.
\newblock \url{https://github.com/Ericsson/eva}.
\newblock (Accessed on 10/26/2020).

\bibitem{feichtenhofer2017detect}
C.~Feichtenhofer, A.~Pinz, and A.~Zisserman.
\newblock Detect to track and track to detect.
\newblock In {\em Proceedings of the IEEE International Conference on Computer
  Vision (ICCV)}, pages 3038--3046, 2017.

\bibitem{Sattar2019ICRA-Fulton-Trash}
M.~{Fulton}, J.~{Hong}, M.~J. {Islam}, and J.~{Sattar}.
\newblock {Robotic Detection of Marine Litter Using Deep Visual Detection
  Models}.
\newblock In {\em 2019 International Conference on Robotics and Automation
  (ICRA)}, pages 5752--5758, May 2019.

\bibitem{girshick2015fast}
R.~Girshick.
\newblock Fast {R-CNN}.
\newblock In {\em Proceedings of the IEEE International Conference on Computer
  Vision (ICCV)}, pages 1440--1448, 2015.

\bibitem{girshick2014rich}
R.~Girshick, J.~Donahue, T.~Darrell, and J.~Malik.
\newblock Rich feature hierarchies for accurate object detection and semantic
  segmentation.
\newblock In {\em Proceedings of the IEEE Conference on Computer Vision and
  Pattern Recognition (CVPR)}, pages 580--587, 2014.

\bibitem{gomez2019caddy}
A.~Gomez~Chavez, A.~Ranieri, D.~Chiarella, E.~Zereik, A.~Babi{\'c}, and
  A.~Birk.
\newblock Caddy underwater stereo-vision dataset for human--robot interaction
  (hri) in the context of diver activities.
\newblock {\em Journal of Marine Science and Engineering}, 7(1):16, 2019.

\bibitem{goodfellow2016deep}
I.~Goodfellow, Y.~Bengio, and A.~Courville.
\newblock {\em Deep learning}.
\newblock MIT press, 2016.

\bibitem{gu2019using}
K.~Gu, B.~Yang, J.~Ngiam, Q.~Le, and J.~Shlens.
\newblock Using videos to evaluate image model robustness.
\newblock {\em arXiv preprint arXiv:1904.10076}, 2019.

\bibitem{he2017mask}
K.~He, G.~Gkioxari, P.~Doll{\'a}r, and R.~Girshick.
\newblock Mask {R-CNN}.
\newblock In {\em Proceedings of the IEEE International Conference on Computer
  Vision (ICCV)}, pages 2961--2969, 2017.

\bibitem{henriques_2012_kcf}
J.~a.~F. Henriques, R.~Caseiro, P.~Martins, and J.~Batista.
\newblock Exploiting the circulant structure of tracking-by-detection with
  kernels.
\newblock In {\em Proceedings of the 12th European Conference on Computer
  Vision (ECCV)}, page 702–715, Berlin, Heidelberg, 2012.

\bibitem{howard2017mobilenets}
A.~G. Howard, M.~Zhu, B.~Chen, D.~Kalenichenko, W.~Wang, T.~Weyand,
  M.~Andreetto, and H.~Adam.
\newblock Mobilenets: Efficient convolutional neural networks for mobile vision
  applications.
\newblock {\em arXiv preprint arXiv:1704.04861}, 2017.

\bibitem{huang2017speed}
J.~Huang, V.~Rathod, C.~Sun, M.~Zhu, A.~Korattikara, A.~Fathi, I.~Fischer,
  Z.~Wojna, Y.~Song, S.~Guadarrama, et~al.
\newblock Speed/accuracy trade-offs for modern convolutional object detectors.
\newblock In {\em Proceedings of the IEEE Conference on Computer Vision and
  Pattern Recognition (CVPR)}, pages 7310--7311, 2017.

\bibitem{islam_diver_2019}
M.~J. {Islam}, M.~{Fulton}, and J.~{Sattar}.
\newblock Toward a generic diver-following algorithm: Balancing robustness and
  efficiency in deep visual detection.
\newblock {\em IEEE Robotics and Automation Letters}, 4(1):113--120, 2019.

\bibitem{jiao2019survey}
L.~Jiao, F.~Zhang, F.~Liu, S.~Yang, L.~Li, Z.~Feng, and R.~Qu.
\newblock A survey of deep learning-based object detection.
\newblock {\em IEEE Access}, 7:128837--128868, 2019.

\bibitem{liu2018mobile}
M.~Liu and M.~Zhu.
\newblock Mobile video object detection with temporally-aware feature maps.
\newblock In {\em Proceedings of the IEEE Conference on Computer Vision and
  Pattern Recognition (CVPR)}, pages 5686--5695, 2018.

\bibitem{liu2019looking}
M.~Liu, M.~Zhu, M.~White, Y.~Li, and D.~Kalenichenko.
\newblock Looking fast and slow: Memory-guided mobile video object detection.
\newblock {\em arXiv preprint arXiv:1903.10172}, 2019.

\bibitem{liu2016ssd}
W.~Liu, D.~Anguelov, D.~Erhan, C.~Szegedy, S.~Reed, C.-Y. Fu, and A.~C. Berg.
\newblock {SSD: S}ingle shot multibox detector.
\newblock In {\em European Conference on Computer Vision (ECCV)}, pages 21--37.
  Springer, 2016.

\bibitem{manfredi2020shift}
M.~Manfredi and Y.~Wang.
\newblock Shift equivariance in object detection.
\newblock {\em arXiv preprint arXiv:2008.05787}, 2020.

\bibitem{mivskovic2016caddy}
N.~Mi{\v{s}}kovi{\'c}, M.~Bibuli, A.~Birk, M.~Caccia, M.~Egi, K.~Grammer,
  A.~Marroni, J.~Neasham, A.~Pascoal, A.~Vasilijevi{\'c}, et~al.
\newblock Caddy—cognitive autonomous diving buddy: Two years of underwater
  human-robot interaction.
\newblock {\em Marine Technology Society Journal}, 50(4):54--66, 2016.

\bibitem{ModasshirICRA2020}
M.~Modasshir and I.~Rekleitis.
\newblock Augmenting coral reef monitoring with an enhanced detection system.
\newblock In {\em IEEE International Conference on Robotics and Automation},
  pages 1874--1880, Paris, France, 2020.

\bibitem{redmon2016you}
J.~Redmon, S.~Divvala, R.~Girshick, and A.~Farhadi.
\newblock You only look once: Unified, real-time object detection.
\newblock In {\em Proceedings of the IEEE Conference on Computer Vision and
  Pattern Recognition (CVPR)}, pages 779--788, 2016.

\bibitem{redmon_2016_yolo}
J.~{Redmon}, S.~{Divvala}, R.~{Girshick}, and A.~{Farhadi}.
\newblock You only look once: Unified, real-time object detection.
\newblock In {\em 2016 IEEE Conference on Computer Vision and Pattern
  Recognition (CVPR)}, pages 779--788, 2016.

\bibitem{redmon_2017_yolov2}
J.~{Redmon} and A.~{Farhadi}.
\newblock Yolo9000: Better, faster, stronger.
\newblock In {\em 2017 IEEE Conference on Computer Vision and Pattern
  Recognition (CVPR)}, pages 6517--6525, 2017.

\bibitem{ren2015faster}
S.~Ren, K.~He, R.~Girshick, and J.~Sun.
\newblock Faster {R-CNN}: Towards real-time object detection with region
  proposal networks.
\newblock In {\em Advances in Neural Information Processing Systems (NIPS)},
  pages 91--99, 2015.

\bibitem{russakovsky2015imagenet}
O.~Russakovsky, J.~Deng, H.~Su, J.~Krause, S.~Satheesh, S.~Ma, Z.~Huang,
  A.~Karpathy, A.~Khosla, M.~Bernstein, et~al.
\newblock Imagenet large scale visual recognition challenge.
\newblock {\em International Journal of Computer Vision}, 115(3):211--252,
  2015.

\bibitem{shankar2019image}
V.~Shankar, A.~Dave, R.~Roelofs, D.~Ramanan, B.~Recht, and L.~Schmidt.
\newblock Do image classifiers generalize across time?
\newblock {\em arXiv preprint arXiv:1906.02168}, 2019.

\bibitem{sunderhauf2018limits}
N.~S{\"u}nderhauf, O.~Brock, W.~Scheirer, R.~Hadsell, D.~Fox, J.~Leitner,
  B.~Upcroft, P.~Abbeel, W.~Burgard, M.~Milford, et~al.
\newblock The limits and potentials of deep learning for robotics.
\newblock {\em The International Journal of Robotics Research},
  37(4-5):405--420, 2018.

\bibitem{szegedy2017inception}
C.~Szegedy, S.~Ioffe, V.~Vanhoucke, and A.~A. Alemi.
\newblock Inception-v4, inception-resnet and the impact of residual connections
  on learning.
\newblock In {\em Proceedings of the Thirty-First AAAI Conference on Artificial
  Intelligence}, pages 4278--4284, 2017.

\bibitem{taori2020measuring}
R.~Taori, A.~Dave, V.~Shankar, N.~Carlini, B.~Recht, and L.~Schmidt.
\newblock Measuring robustness to natural distribution shifts in image
  classification.
\newblock {\em arXiv preprint arXiv:2007.00644}, 2020.

\bibitem{tf_example}
{TensorFlow}.
\newblock {\em tf.train.Example}, 2020.
\newblock {\small \url{https://bit.ly/3jO6tuR}}. Accessed 10-31-2020.

\bibitem{tf_sequence}
{TensorFlow}.
\newblock {\em {tf.train.SequenceExample}}, 2020.
\newblock {\small \url{https://bit.ly/3oMDdYX}}. Accessed 10-31-2020.

\bibitem{zhang2017stability}
H.~Zhang and N.~Wang.
\newblock On the stability of video detection and tracking.
\newblock {\em arXiv preprint arXiv:1903.10172}, 2017.

\bibitem{zhao2019object}
Z.-Q. Zhao, P.~Zheng, S.-t. Xu, and X.~Wu.
\newblock Object detection with deep learning: A review.
\newblock {\em IEEE Transactions on Neural Networks and Learning Systems},
  30(11):3212--3232, 2019.

\bibitem{zhu2018towards}
X.~Zhu, J.~Dai, L.~Yuan, and Y.~Wei.
\newblock Towards high performance video object detection.
\newblock In {\em Proceedings of the IEEE Conference on Computer Vision and
  Pattern Recognition (CVPR)}, pages 7210--7218, 2018.

\bibitem{zhu2017flow}
X.~Zhu, Y.~Wang, J.~Dai, L.~Yuan, and Y.~Wei.
\newblock Flow-guided feature aggregation for video object detection.
\newblock In {\em Proceedings of the IEEE International Conference on Computer
  Vision (ICCV)}, pages 408--417, 2017.

\end{thebibliography}

\end{document}